\documentclass[11pt]{article}
\usepackage[final]{acl}
\usepackage{times}
\usepackage{latexsym}
\usepackage[T1]{fontenc}
\usepackage[utf8]{inputenc}
\usepackage{microtype}
\usepackage{inconsolata}
\usepackage{graphicx}
\usepackage{newunicodechar}
\usepackage{tipa}
\newunicodechar{Ṛ}{\text{\dj}}

\title{Accent Placement Models for Rigvedic Sanskrit Text}

\author{Akhil Rajeev P \and Annarao Kulkarni \\
  Indian Heritage Language Computing Team \\
  Special and Strategic Projects (SSP) Group \\
  Centre for Development of Advanced Computing (C DAC), Bangalore \\
  \texttt{akhilrajeev@cdac.in} \quad \texttt{kulkarni@cdac.in}
}

\begin{document}
\maketitle
\begin{abstract}
The Rigveda, among the oldest Indian texts in Vedic Sanskrit, employs a distinctive pitch-accent system : udātta, anudātta, svarita whose marks encode melodic and interpretive cues but are often absent from modern e-texts. This work develops a parallel corpus of accented-unaccented ślokas and conducts a controlled comparison of three strategies for automatic accent placement in Rigvedic verse: (i) full fine-tuning of ByT5, a byte-level Transformer that operates directly on Unicode combining marks, (ii) a from-scratch BiLSTM-CRF sequence-labeling baseline, and (iii) LoRA-based parameter-efficient fine-tuning atop ByT5.

Evaluation uses Word Error Rate (WER) and Character Error Rate (CER) for orthographic fidelity, plus a task-specific Diacritic Error Rate (DER) that isolates accent edits. Full ByT5 fine-tuning attains the lowest error across all metrics; LoRA offers strong efficiency-accuracy trade-offs, and BiLSTM-CRF serves as a transparent baseline. The study underscores practical requirements for accent restoration - Unicode-safe preprocessing, mark-aware tokenization, and evaluation that separates grapheme from accent errors - and positions heritage-language technology as an emerging NLP area connecting computational modeling with philological and pedagogical aims. Results establish reproducible baselines for Rigvedic accent restoration and provide guidance for downstream tasks such as accent-aware OCR, ASR/chant synthesis, and digital scholarship.
\end{abstract}

\section{Introduction}

The \textit{Rigveda}, an ancient collection of \emph{\d{r}k}-s (hymns) composed in Vedic Sanskrit, encodes its recitational tradition through a sophisticated accent system. Words in Vedic Sanskrit bear accented syllables, and each Veda has its own set of accent markers, with some shared across Vedic corpora. The detailed list of accent markers used in the Vedas is standardized in ISO/ISCII\_Ammex-G.\footnote{For descriptive overviews see \citet{ISCII_AnnexG}; for phonetic discussion of the independent \emph{svarita}, see \citet{begus2016svarita}.}

\noindent Rigvedic phonology distinguishes three tones: \textbf{Udātta} (high tone, normally unmarked), \textbf{Anudātta} (low tone, shown by a mark below the character; U+0952), and \textbf{Svarita} (rising-falling tone, marked above the character; U+0951). These accent signs guide chanting and preserve tonal precision in oral tradition.

\noindent These markers prescribe tone or pitch for recitation, and are also embedded semantic units: a change in accent can alter the meaning of a word. The phonetic rules governing accents are described in the \emph{Pr\=ati\'s\=akhya}-s and \emph{\'Sik\d{s}\=a\'s\=astra} texts, with the \emph{\d{R}gveda Pr\=ati\'s\=akhya} serving as the authoritative source for Rigvedic phonology. The \emph{Nigha\d{n}\d{t}u} provides a lexicon of Vedic words, and fully appreciating why a syllable bears a particular accent typically requires expertise in Sanskrit grammar, \emph{chandas} (metrics), \emph{nirukta} (etymology), and phonetics.

\noindent Despite its linguistic centrality, many searchable e-texts and NLP resources omit accents due to encoding limitations or design choices prioritizing searchability \citep{unicodeDeva17,unicodeVedic17,accentsCologne}. This omission hampers philological research, chanting pedagogy, and speech systems that depend on tonal cues \citep{hellwig2020vedicTB,vedavani2025}. Accent distinctions are essential for oral instruction, yet learners using unaccented corpora cannot reconstruct melodic contours. In speech technology, ASR or TTS systems trained on unaccented data fail to capture prosody vital for faithful recitation. Automatic accent restoration therefore represents both a \textbf{technical challenge}-a low-resource sequence labeling task on metrical Sanskrit verse with pervasive sandhi-and a \textbf{cultural-heritage challenge} vital to preserving an oral tradition.

\noindent Recent NLP advances make such restoration feasible. Byte- and character-level transformers restore diacritic-like markers without brittle tokenization \citep{xu2022byt5}, and parameter-efficient fine-tuning lowers adaptation cost for niche low-resource domains \citep{houlsby2019parameter,hu2022lora}. Parallel work on Arabic, Hebrew, and Yorùbá shows the effectiveness of normalization-aware pipelines and diacritic-sensitive architectures \citep{alqahtani2020acl,gershuni2022nakdimon,rosenthal2024dnikud,menakbert2024,oyad2024}, suggesting methodological transferability even though Vedic accent remains unexplored.

\noindent This study investigates whether modern AI models can automatically accent unaccented Rigvedic hymns without expert linguistic rules. We construct a parallel corpus of accented-unaccented verse pairs and evaluate three strategies: (i) full ByT5 fine-tuning \citep{xu2022byt5}, (ii) a BiLSTM-CRF sequence labeler \citep{huang2015bidirectional}, and (iii) LoRA-based ByT5 tuning \citep{hu2022lora}. Evaluation using Word, Character, and Diacritic Error Rates (WER, CER, DER) shows full ByT5 achieves the lowest errors, LoRA balances accuracy and efficiency, and BiLSTM-CRF provides a reproducible baseline. Our released corpus and code aim to advance accent-aware OCR, ASR, and pedagogy for Vedic studies \citep{tsukagoshi2025ocr,vedavani2025}.

\section{Related Work}
\paragraph{Computational Sanskrit and Vedic resources:}
Sanskrit NLP has focused on segmentation, sandhi splitting, morphology, and syntax - foundations for accent restoration. Early tools such as \textsc{SanskritTagger} and sentence boundary detectors processed punctuation-light text \citep{hellwig2010,hellwig2016sbd}. Neural methods advanced segmentation via character-CNN/LSTM and graph inference \citep{hellwig2018seg,krishna2016pcs}. The Sanskrit Heritage platform, distributed processing stacks, and the UD-style \emph{Vedic} Treebank supply lexical and syntactic supervision useful for accent diagnostics \citep{goyal2012sh,huetHeritage,hellwig2020vedicTB}. Indian research further formalized dependency relations for Sanskrit grammar \citep{kulkarni2020dep}. Recent work introduces accent-aware OCR and ASR benchmarks for Vedic Devanagari, defining a new context for restoration \citep{tsukagoshi2025ocr,vedavani2025}.

\paragraph{Standardization of Vedic characters:}
The extended character repertoire for Vedic scripts defined in Annex~G of \citet{ISCII_AnnexG} (ISCII) provided the initial framework for digital representation of Vedic accents and combining marks. This early standard, based on an 8-bit encoding scheme, served as the foundation for later Unicode integration. Its specifications were incorporated into the Unicode \emph{Vedic Extensions} block (U+1CD0-U+1CFX), ensuring compatibility with Devanagari and related Brahmic scripts and enabling cross-platform rendering of accents and tonal signs. This standardization has been central to developing searchable, accent-preserving corpora and tools for computational Vedic studies \citep{unicodeDeva17,unicodeVedic17}.

\paragraph{Diacritics and accents beyond Sanskrit:}
Arabic, Hebrew, and Yorùbá diacritic restoration offer methodological parallels. Accuracy correlates with modeling capacity and domain adaptation, from multitask setups to "diacritics-in-the-wild" corpora \citep{alqahtani2020acl,elgamal2024wild}. Hebrew and Yorùbá studies use compact character LSTMs or transformer variants (NAKDIMON, MenakBERT, D-Nikud, T5) \citep{gershuni2022nakdimon,menakbert2024,rosenthal2024dnikud,oyad2024}.

\paragraph{Modeling choices:}
Byte and character-level transformers avoid fragile tokenization in diacritic-rich scripts; the T5 variant used here is ByT5-Sanskrit
(\citep{nehrdich-etal-2024-one}), a byte-level model trained specifically for Sanskrit NLP tasks, which avoids fragile tokenization in diacritic-rich scripts and performs strongly on UTF-8 text. Parameter-efficient transfer (adapters, LoRA) lowers cost for low-resource tasks such as Rigvedic accenting
\citep{houlsby2019parameter,hu2022lora}. BiLSTM-CRF remains a transparent baseline for sequence labeling \citep{huang2015bidirectional}.

\section{Dataset}

An in-house, validated \textit{Rigveda} corpus developed at C-DAC is used, comprising 10{,}552 hymns organized into 10 \emph{maṇḍalas} and 1{,}028 \emph{sūktas}. From this resource, a parallel corpus of 22{,}740 aligned verse pairs is constructed: each entry pairs an unaccented verse with its diacritically marked counterpart for supervised training and evaluation.\footnote{The C-DAC Rigveda parallel corpus will be made available through the Indian Knowledge Base platform at \url{https://indianknowledgebase.in/}.}

\noindent Provenance fields (maṇḍala-sūkta-ṛc identifiers) accompany each record.

\noindent\textbf{Example:}\\
\textbf{Unaccented}\\
\includegraphics[width=0.83\linewidth]{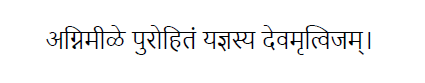}

\noindent\textbf{Accented}\\
\includegraphics[width=0.8\linewidth]{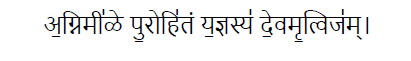}

\noindent The accented form adds pitch cues via combining marks while core graphemes remain unchanged - motivating CER for orthography and DER for accent-specific edits.

\paragraph{Splits:}
Train / test / dev (validation) partitions are drawn from the in-house Rigveda corpus, with stratification by maṇḍala and sūkta length to mitigate topical leakage. From the total of 22,740 aligned verse pairs, we adopt the train / validation / test split, as shown in Table \ref{tab:split}.

\begin{table}[ht]
\centering
\caption{\textbf{Train / development / test split of the aligned corpus (22,740 verse pairs).}}
\label{tab:split}
\small
\renewcommand{\arraystretch}{1.2} 
\begin{tabular}{lrr}
\hline
\textbf{Split} & \textbf{Verse pairs} & \textbf{Percentage} \\
\hline
\textbf{Train}        & \textbf{19,329} & \textbf{85\%} \\
\textbf{Development}  & \textbf{2,274}  & \textbf{10\%} \\
\textbf{Test}         & \textbf{1,137}  & \textbf{5\%} \\
\hline
\textbf{Total}        & \textbf{22,740} & \textbf{100\%} \\
\hline
\end{tabular}
\end{table}







\section{Methodology}
We evaluate three models:  

\begin{enumerate}
    \item \textbf{Full Fine-tuning (ByT5):} The multilingual ByT5 model was fine-tuned end-to-end. We used learning rate $3e{-5}$, batch size 32, and trained for 10 epochs.  
    \item \textbf{BiLSTM-CRF:} This model used 256-d embeddings, a 2-layer BiLSTM (hidden size 512), and a CRF decoding layer. Dropout 0.3 was applied. Training used Adam ($lr=1e{-3}$) for 20 epochs.  
    \item \textbf{LoRA Fine-tuning (ByT5):} LoRA with rank 8 and $\alpha=16$ was applied to the self-attention projection matrices. Only 0.5\% of parameters were updated.
\end{enumerate}

\section{Evaluation Metrics}

System performance is assessed using three complementary string-level metrics designed to capture lexical, orthographic, and diacritic-specific accuracy.  

\begin{itemize}
    \item \textbf{Word Error Rate (WER)} measures \textbf{token-level edit distance} (Insertions, Deletions, Substitutions), reflecting overall lexical fidelity. It is normalized by the number of reference tokens.
    \item \textbf{Character Error Rate (CER)} computes \textbf{character-level edit distance}, excluding whitespace. This metric is sensitive to fine-grained orthographic deviations, making it suitable for morphologically rich Sanskrit text.
    \item \textbf{Diacritic Error Rate (DER)} isolates errors in \textbf{accent symbols (diacritics)} alone, disregarding base characters. It quantifies the precision of accent placement (tonal correctness), normalized over the total diacritic instances in the reference.
\end{itemize}

\noindent Together, these metrics provide complementary views of model behavior: WER captures global token accuracy, CER measures character integrity, and DER specifically reflects accent restoration performance at the sub-character level.

\section{Results and Discussion}
\begin{table}[h]
\centering
\begin{tabular}{lccc}
\hline
\textbf{Method} & \textbf{WER} & \textbf{CER} & \textbf{DER} \\
\hline
Full FT (ByT5) & \textbf{0.1023} & \textbf{0.0246} & \textbf{0.0685} \\
BiLSTM--CRF & 0.2367 & 0.0448 & 0.3197 \\
LoRA FT (ByT5) & 0.3614 & 0.1042 & 0.1598 \\
\hline
\end{tabular}
\caption{Performance of Sanskrit accent placement models. Best scores in bold.}
\label{tab:results}
\end{table}

\noindent Our findings highlight three insights:  
\paragraph{Transformer advantage:} Full ByT5 fine-tuning outperforms alternatives by a wide margin, confirming that large pretrained models adapt well even to heritage tasks with limited training data.  

\paragraph{Diacritic modeling challenge:} BiLSTM--CRF achieves tolerable WER and CER but fails dramatically in DER. This suggests that traditional models cannot capture pitch diacritic patterns without explicit linguistic priors.  

\paragraph{Efficiency vs. fidelity:} LoRA reduces trainable parameters by orders of magnitude but suffers in WER/CER. Interestingly, its DER surpasses BiLSTM--CRF, indicating that localized diacritic learning may be partially preserved.  

\noindent Beyond metrics, these results matter for applications: chanting synthesis requires low DER, while digital philology may tolerate higher CER if semantic accents are preserved.  

\section{Error Analysis}
To examine model behavior beyond aggregate metrics, we manually analyzed 200 mispredicted verses from the test set, comparing error tendencies across ByT5 (full fine-tuning), BiLSTM–CRF, and ByT5-LoRA. Four categories emerged: accent misplacement, omission or over-generation, accent-type confusion, and boundary errors.

\paragraph{Accent Misplacement}:The dominant category (46.8\%) involved accents shifted by one mora within the correct syllabic span (e.g., \textit{deva\textsubbar{m}ṛtvijam} → \textit{devamṛ\textsubbar{t}vijam}). ByT5 had the lowest misplacement rate (18.2\%), while BiLSTM–CRF (41.5\%) and LoRA (33.4\%) showed weaker morphemic control, suggesting that ByT5’s byte-level encoding captures compound co-occurrence patterns, whereas LoRA underfits longer phonological sequences.

\paragraph{Omission and Over-generation:}These formed 26.3\% of all errors, mostly in verses with multiple enclitic particles (\textit{ha, ca, u}). BiLSTM–CRF tended to over-generate (14.8\%), while LoRA favored omission (11.2\%), reflecting a conservative decoding bias from low-rank adaptation.

\paragraph{Accent-Type Confusion:}About 15.1\%) involved udātta–svarita swaps, common in reduplicated or rhythmic verb forms, indicating a need for explicit tone hierarchy modeling.

\paragraph{Boundary and Tokenization Errors:}A smaller share (8.7\%)) arose from accent drift across pāda or punctuation boundaries. BiLSTM–CRF was most affected due to fixed segmentation, whereas ByT5’s byte-level representation mitigated drift.

\paragraph{Cross-Metric Correlation:}Diacritic Error Rate (DER) correlated strongly with Character Error Rate (CER) ($r$ = 0.82) but weakly with Word Error Rate (WER) ($r$ = 0.39), confirming accent restoration as a sub-character orthographic task.

\paragraph{Qualitative Observations:}Mid-frequency stems (\textit{agní}, \textit{soma}, \textit{indra}) were accented correctly across models, while rare words like \textit{puruṣṭuta} showed erratic realizations.

\section{Conclusion}
We introduced the first benchmark for automatic accent restoration in Rigvedic Sanskrit, evaluated with \textit{Word Error Rate (WER)}, \textit{Character Error Rate (CER)}, and the task-specific \textit{Diacritic Error Rate (DER)} focused on accent deviations. Full fine-tuning of ByT5 achieves the strongest results, LoRA balances efficiency and accuracy, and BiLSTM--CRF serves as a transparent baseline. This work demonstrates that modern NLP methods can be effectively adapted to heritage-language processing despite data sparsity and domain constraints.
\noindent Beyond restoration accuracy, the framework holds promise for enabling systematic \textit{prosodic annotation} of Vedic corpora, thereby facilitating deeper linguistic and chanting analyses. Its interpretability could further support \textit{explainable AI} approaches to modeling oral–textual correspondences. 

\section{Acknowledgments}
    We acknowledge Dr. Janaki C. H. and Dr. S. D. Sudarsan for their guidance and administrative facilitation throughout this work. Additionally,our sincere thanks go to Sinchana Bhat and Archana Bhat, Classical Sanskrit Linguists at C-DAC Bangalore, for their crucial contributions in data annotation and human evaluation.

\section{Limitations}
Our study has the following limitations:

\begin{enumerate}
    \item \textbf{Data size:} The corpus is relatively small compared to modern NLP benchmarks, restricting model generalization and robustness.
    \item \textbf{Evaluation metrics:} We use WER, CER, and DER, which measure surface accuracy but do not capture alignment with deeper metrical or phonological rules described in traditional sources.
    \item \textbf{Model coverage:} We evaluate three approaches (ByT5, LoRA, BiLSTM--CRF). Other architectures, such as non-autoregressive transformers, graph-based methods, or phonology-aware encoders, are not explored.
\end{enumerate}

\bibliography{custom}  


\end{document}